\documentclass[preprint,12pt]{elsarticle}

\usepackage{amssymb}
\usepackage{amsmath}
\usepackage{lineno}
\usepackage{booktabs} 
\usepackage{tabularx}
\usepackage{float}  
\usepackage{stfloats}
\usepackage{diagbox}
\usepackage{comment}
\usepackage{diagbox}
\usepackage{subcaption}
\usepackage{hyperref}

\makeatletter
\def\ps@pprintTitle{%
  \let\@oddhead\@empty
  \let\@evenhead\@empty
  \let\@oddfoot\@empty
  \let\@evenfoot\@empty
}
\makeatother
\begin{document}

\begin{frontmatter}

\title{Riverine Land Cover Mapping through Semantic Segmentation of
Multispectral Point Clouds} %% Article title

\author[FGI,UTUT]{Sopitta Thurachen} %% Author name
\author[FGI]{Josef Taher} %% Author name 
\author[FGI]{Matti Lehtomäki}
\author[FGI]{Leena Matikainen}
\author[UTU]{Linnea Blåfield}
\author[UTU,UM]{Mikel Calle Navarro}
\author[FGI]{Antero Kukko}
\author[UTUT]{Tomi Westerlund}
\author[FGI]{Harri Kaartinen}

%% Author affiliation
\affiliation[FGI]{organization={Department of Remote Sensing and Photogrammetry, Finnish Geospatial Research Institute FGI},%Department and Organization
            addressline={Vuorimiehentie 5}, 
            city={Espoo},
            postcode={02150}, 
            %state={},
            country={Finland}}
            
\affiliation[UTUT]{organization={Faculty of Technology, University of Turku},%Department and Organization
            %addressline={Henrikinkatu 2}, 
            city={Turku},
            postcode={20014}, 
            %state={},
            country={Finland}}   
            
\affiliation[UTU]{organization={Department of Geography and Geology, University of Turku},
            city={Turku},
            postcode={20014}, 
            %state={},
            country={Finland}}

\affiliation[UM]{organization={Faculty of Geological Sciences, Complutense University of Madrid},%Department and Organization
            addressline={Avda. de Séneca, 2 Ciudad Universitaria}, 
            city={Madrid},
            postcode={28040}, 
            %state={},
            country={Spain}}

%% Abstract
\begin{abstract}
Accurate land cover mapping in riverine environments is essential for effective river management, ecological understanding, and geomorphic change monitoring. This study explores the use of Point Transformer v2 (PTv2), an advanced deep neural network architecture designed for point cloud data, for land cover mapping through semantic segmentation of multispectral LiDAR data in real-world riverine environments. We utilize the geometric and spectral information from the 3-channel LiDAR point cloud to map land cover classes, including sand, gravel, low vegetation, high vegetation, forest floor, and water. The PTv2 model was trained and evaluated on point cloud data from the Oulanka river in northern Finland using both geometry and spectral features. To improve the model's generalization in new riverine environments, we additionally investigate multi-dataset training that adds sparsely annotated data from an additional river dataset. Results demonstrated that using the full-feature configuration resulted in performance with a mean Intersection over Union (mIoU) of 0.950, significantly outperforming the geometry baseline. Other ablation studies revealed that intensity and reflectance features were the key for accurate land cover mapping. The multi-dataset training experiment showed improved generalization performance, suggesting potential for developing more robust models despite limited high-quality annotated data. Our work demonstrates the potential of applying transformer-based architectures to multispectral point clouds in riverine environments. The approach offers new capabilities for monitoring sediment transport and other river management applications.
\end{abstract}

%%Graphical abstract
%\begin{graphicalabstract}
%\includegraphics{grabs}
%\end{graphicalabstract}

%%Research highlights
%\begin{highlights}
%\item Research highlight 1
%\item Research highlight 2
%\end{highlights}

%% Keywords

\end{frontmatter}

%% main text
%%
\section{Introduction}

Riverine environments are highly dynamic and ecologically significant landscapes, where the interplay of physical processes (such as fluvial processes and sediment dynamics) and land cover patterns shapes their form and functionality \citep{snelder2002multiscale}. For example, watershed land cover affects water flow, water chemistry, and the type of sediment in stream channels \citep{snelder2002multiscale}. The riverbed sediments and materials are essential for defining aquatic habitat conditions, as the riverbed materials serve as feeding, spawning, and sheltering grounds for aquatic organisms \citep{irie2023classification}. In addition, the establishment of vegetation on certain sediment deposits can in turn modify flow patterns and promote further sedimentation, resulting in further altering the river’s morphology \citep{irie2023classification}.
Considering the interconnected nature of these ecological and hydrological processes, accurate spatial information obtained through river mapping is essential for effective river management, ecological understanding, and geomorphic change. 
However, traditional mapping methods have certain limitations. Field-based surveys can be labor-intensive and often impractical to scale up \citep{irie2023classification}. Traditional two-dimensional (2D) remote sensing approaches such as aerial or satellite imagery can map larger areas, but these methods also have their own setbacks in complex riverine environments. Passive 2D optical remote sensing, for example, can have limitations in penetrating  water bodies (depending on factors such as water visibility, depth, and  water column attenuation \citep{legleiter2019remote}) and in dense vegetation canopies to reveal understoreys \citep{akasheh2008detailed,latifi2016estimating}.
Furthermore, overhead 2D imagery lacks the vertical dimension of terrain structure, making it difficult to distinguish certain land cover types. These limitations of conventional methods hinder their capacity in capturing complex and dynamic three-dimensional (3D) representations of river environments in a timely and accurate manner.

New developments in remote sensing technology, especially multispectral laser scanning \citep{matikainen2017object,morsy2017multispectral,takhtkeshha2024multispectral}, have arisen as a promising tool to address these limitations. Multispectral airborne laser scanning (ALS, also LiDAR) systems have enabled acquisition of not only 3D point clouds but also spectral information of the mapped objects. This enables more accurate land cover mapping compared to a single wavelength system, which lacks spectral information for classifying similar surfaces in complex environments \citep{ma2019multispectral}.

The Finnish Geospatial Research Institute (FGI) has developed a helicopter-borne triple-wavelength (TW) airborne laser scanning system called HeliALS-TW, enabling the simultaneous capture of high-resolution 3D point clouds along with spectral information at each wavelength. The HeliALS-TW system combines three RIEGL scanners, VUX-1HA, miniVUX-3UAV, and VQ-840-G, operating at wavelengths 1550\,nm, 905\,nm, and 532\,nm, respectively. The system has been used to map river environments for this study. 

Previously, many researchers have explored multispectral ALS in the context of land cover classification \citep{wang2014airborne,wichmann2015evaluating,matikainen2017object,ma2019multispectral,ghaseminik2021land,shi2021land}. Some studies utilized multispectral point clouds from the Optech Titan system (Teledyne Optech, Ontario, Canada) with classical classifiers such as Random Forest (RF) and Support Vector Machine (SVM) \citep{matikainen2017object,shi2021land}. \citet{ghaseminik2021land} studied a specific case of classifying multispectral point clouds using RF with data imbalance. \citet{ma2019multispectral} used convolutional neural networks (CNN)  for land cover classification. These studies have demonstrated the effectiveness of using multispectral ALS data for land cover classification tasks.

In addition, technological advancements in remote sensing have also led to an increase in both the quantity and quality of different types of geospatial data, e.g., satellite imagery and LiDAR data giving rise to data-driven approaches such as deep learning. Deep learning (DL) for point cloud classification and segmentation has progressed quickly, in the field of remote sensing and land cover mapping. Traditional classification methods for both 2D and 3D data usually involve feature engineering, and features are then fed into classifiers such as RF and SVM. Deep Learning, especially well-established convolutional neural networks and, more recently, transformer-based architectures, can learn task-specific features from spatial data without having to manually engineer them. Transformer-based models, known for their attention mechanism and the capability to capture global contexts, have recently demonstrated significant potential in 3D point cloud space \citep{zhao2021point,yang2025swin3d}. 

The Point Transformer model~\citep{zhao2021point} was one of the first to successfully utilize a self-attention mechanism on 3D point clouds on several benchmark datasets such as S3DIS~\citep{armeni20163d}, ModelNet40~\citep{wu20153d}, and ShapeNet~\citep{chang2015shapenet}. The Point Transformer v2 (PTv2)~\citep{wu2022point} is the improved version of the original Point Transformer \citep{zhao2021point}. PTv2 has grouped vector attention and partition-based pooling. This version shows an increased performance on the benchmark datasets mentioned above. While transformer-based architectures have illustrated promising performance on those benchmark datasets, their applicability to multispectral point cloud data, especially in riverine environments, still has room for exploration.

Furthermore, additional challenges are introduced when applying advanced deep learning models, including newly developed transformer-based architectures, on natural multispectral point clouds such as those from riverine environments. One major challenge is class imbalance in the training data. River corridors can have dominant land cover classes (e.g., vegetation), while other essential classes like sediments are not well represented in the data. Another challenge is cross-area generalization. Environmental conditions can differ substantially between river sites, for example in the terrain types and in the distribution of land cover classes. These challenges emphasize the importance of evaluating whether advanced models like transformer-based architectures can be effectively adapted to real-world environmental point cloud, and what strategies can potentially improve their robustness.

The main focus of this research is to evaluate the effectiveness of a transformer-based architecture for land cover mapping in riverine environments, using multispectral point cloud data with an emphasis on handling class imbalance and enhancing cross-area generalization with a multi-dataset training framework. To this end, we selected the network PTv2 due to its strong performance on established 3D point cloud datasets, its efficient grouped vector attention and partition-based pooling, and its utilization of point clouds as inherently unordered structures. We aim to tackle the following objectives:

\begin{itemize}
    \item Using multispectral point clouds and semantic segmentation to evaluate the effectiveness of PTv2 for land cover mapping in riverine environment segmentation.
    \item Adding spectral features to investigate whether they can improve land cover classification performance compared to results with geometry alone.
    \item Conducting an ablation study on each individual spectral feature: intensity, reflectance, deviation, and amplitude to analyze their individual contributions to land cover classification performance.
    \item Utilizing a multi-dataset training framework with sparsely annotated data to tackle class imbalance and to improve PTv2 generalization capability in new environments.

\end{itemize}

\section{Related Work}

\subsection{Classical Machine Learning Methods for Land Cover Mapping}

Traditional machine learning methods, including RF \citep{breiman2001random} and SVMs \citep{cortes1995support} have greatly contributed to the progress of land cover mapping tasks. The main reasons why RF has gained popularity in remote sensing research are mainly because of its classification accuracy, robustness, and ability to handle high-dimensional data \citep{belgiu2016random}. RF has been applied in different land cover mapping applications, such as urban buildings \citep{belgiu2014quantitative}, biomass \citep{frazier2014characterization}, and tree canopy \citep{karlson2015mapping}, using various data sources such as thermal imagery, hyperspectral imagery, LiDAR and multispectral radar \citep{belgiu2016random}.
SVMs have also demonstrated promising performance in various land cover classification tasks \citep{mountrakis2011support} such as vegetation classification with hyperspectral imagery \citep{gualtieri1999support}, forest species classification by fusing hyperspectral imagery with LiDAR \citep{dalponte2008fusion} and urban area mapping with satellite imagery (SPOT5) \citep{inglada2007automatic}. One of the main advantages of SVMs is their ability to generalize well with limited training data \citep{mountrakis2011support}. These traditional machine learning approaches have established a strong foundation for automated land cover mapping, which has made the development of more advanced methods possible.

\subsection{Deep Learning for Land Cover Mapping}

DL methods have become state-of-the-art approaches in many computer vision tasks including land cover mapping. A major benefit of DL is its ability to learn patterns from raw input data directly without extensive manual feature engineering while still performing well compared to traditional methods \citep{lecun2015deep}.   

In remote sensing, land cover mapping often uses diverse data types, such as aerial photographs~\citep{mboga2020fully,mora2015land}, satellite imagery~\citep{lu2012land,kim2016land,mcroberts2002using}, and LiDAR data~\citep{shi2021land,matikainen2020combining}. DL algorithms have consistently demonstrated robust performance in handling these 2D and 3D data types. 
%good review papers~\cite{belgiu2016random}~\cite{ma2019deep}  

Deep Convolutional Neural Networks (CNNs) are among the most widely used DL techniques in remote sensing data analysis due to their ability to extract spatial relationships from remote sensing data. CNNs have been successfully used for different land cover mapping such as urban land cover classification~\cite{shi2021land}, agricultural land use classification~\cite{simon2022convolutional}, and vegetation type mapping~\cite{liao2020synergistic}.
%[add some more applications here, good review paper : Review on Convolutional Neural Networks (CNN) in vegetation remote sensing]

Beyond CNNs, other DL architectures such as Vision Transformers \citep{dosovitskiy2020image} and Autoencoders (AE) \citep{ballard1987modular} have also been studied for mapping land covers. Vision Transformers captured global contextual information through self-attention mechanisms, resulting in improved classification accuracy \citep{bazi2021vision,bi2022vision}. Autoencoders especially variational autoencoders (VAEs) have demonstrated their advantage in situations where annotated training data is limited. When used in unsupervised and semi-supervised settings, VAEs can extract latent features from unlabeled or partially labeled remote sensing datasets \citep{zhao2022semisupervised}. 

Despite these advances, using DL methods in land cover mapping still has real-world challenges to be addressed. One problem is the limited availability of high-quality annotated data for training DL models. Curating such a dataset is not only expensive but also time-consuming. 

\subsection{Deep Learning for Riverine Land Cover Mapping}

Recent research on river-related land cover mapping has been dominated mainly by 2D satellite imagery data and CNN architectures \citep{gonzales2022geospatial}. \citet{moortgat2022deep} utilized CNNs to classify rivers in the Arctic using satellite imagery from Quickbird-2, WorldView-1, WorldView-2, WorldView-3, and GeoEye satellites. Extending CNNs with attention, \citet{Marjani2024CVTNetAF} introduced CVTnet, a hybrid network between CNN and ViT for wetland mapping from Sentinel-1 and Sentinel-2 data. In alpine environments, \citet{qichi2023novel} demonstrated that deep CNNs maintain performance despite high topography, while \citet{tzepkenlis2023efficient} used an improved U-TAE for land cover classification for Sentinel satellite image time series. 

In addition to 2D imagery data, there are some works relevant to riverine land cover mapping using point cloud data. Because fluvial environments usually possess strong vertical structure, such as vegetation, some researchers have turned to LiDAR. Although some of these works utilized traditional classifiers. \citet{laslier2019mapping} used bispectral LiDAR (532 nm and 1064 nm) and an RF classifier to map riparian vegetation along the Sélune River in France. Their classification focused only on vegetation (riparian indicators such as tree species and strata density) and found that bispectral intensity contributed only marginally to the accuracy because it was not corrected. \citet{gomez2022mapping} mapped subaerial facies in Yuba River in California using topobathymetric airborne LiDAR and RF. This work only focused on different types of sediments, such as large gravel, sand/fine gravel, and large cobble. In addition to using RF for mapping, some works use DL to map river land covers.~\cite{bolick2021evaluation} utilized the U-Net DL model to map land cover and shading along the Chauga River on high-resolution imagery, also using LiDAR to incorporate vegetation structure. While achieving overall accuracies of 95–97\%, LiDAR was not used in the classification itself. \citet{yoshida2022airborne} applied a modified DeepLabV3+ model to aerial photographs together with airborne laser bathymetry (ALS)-derived features for riparian land cover classification along the Asahi River in Japan. The results showed great improvement over their previous unsupervised ALS-based method, though the classification was still performed on 2D imagery.

\subsection{Deep Learning in 3D Point Clouds for Land Cover Mapping}
The application of DL to 3D point cloud data has changed land cover mapping by allowing direct processing of point cloud data. Point clouds contain 3D geometry that is particularly helpful for differentiating the structure of various land cover types. In addition to 3D geometric information, point clouds are often captured with radiometric information, which is also beneficial for land cover mapping.

Point clouds are unstructured. Early approaches to DL with point clouds relied on converting the unstructured point clouds into regular 3D voxel grids or projecting point clouds from 3D to 2D multiview images so that 3D or 2D CNN-based methods could be used. For example, \citet{maturana2015voxnet} introduced VoxNet. In this work, point clouds were represented as volumetric occupancy grids and then 3D CNNs were applied for object recognition. While these approaches could be effective, they struggled with the sparsity and scale variations common in point clouds and often incurred information loss during conversion.

A major breakthrough in DL for point clouds came with the introduction of PointNet \citep{qi2017pointnet}, the first network to directly work with unstructured point sets in 3D by using shared Multilayer Perceptrons (MLPs). PointNet++ \citep{qi2017pointnet++} built on that foundation by introducing a hierarchical neural network that uses PointNet recursively on the point sets in metric spaces. This approach allowed the network to better capture local features at multiscale scales \citep{qi2017pointnet++}.

In geospatial applications, researchers have developed specialized adaptations of these architectures (PointNet, PointNet++) to improve performance. When applied to ALS data, an adaptation of PointNet++ achieved 80.6 percent overall accuracy on the ISPRS 3D Semantic Labeling benchmark \citep{winiwarter2019classification}. Besides using only geometry, Yousefhussien et al.~\cite{yousefhussien2018multi} presented a multimodal network, which is an adaptation of PointNet utilizing spectral data derived from 2D IR-R-G imagery to classify land covers. The network achieved 81.6 percent overall accuracy on the ISPRS 3D Semantic Labeling benchmark dataset.

\subsection{Transformer-Based Method in 3D Point Cloud for Land Cover Mapping}

The success of transformer-based architectures in Computer Vision (CV) and Natural Language Processing (NLP) has led to increased interest in applying these models within the remote sensing community. Transformers are now being used in various applications, including land cover mapping. Transformer networks fundamentally use self-attention \citep{vaswani2017attention} to effectively capture global relationships between input and output. 

In the field of 3D point clouds, transformer architectures have addressed the unique challenges inherent in the unstructured nature of point clouds. The Point Transformer (PT) \citep{zhao2021point} and Point Cloud Transformer (PCT) \citep{guo2021pct} are some of the most established transformer networks for point cloud data processing and analysis. By utilizing the permutation-invariant property, these models can effectively handle the unstructured characteristic of point clouds.

Both PT and PCT utilize the concept of self-attention to analyze 3D point cloud data. Each point in the cloud is treated as a word in natural language. Then self-attention is applied to its neighboring points. This approach allows the capture of complex spatial relationships in 3D space.

Although these two models share this fundamental approach, they are still different in their specific implementations. PT uses vector attention, whereas PCT employs offset attention, which calculates the difference between the input of the self-attention module and the attention features rather than directly using the attention features for the aggregation. 

These developments have shown great results in several 3D understanding tasks, including semantic segmentation, which is crucial for automated land cover mapping.

For geospatial applications, recent research has explored improved transformer models for classifying land cover utilizing multispectral LiDAR point clouds that include multi-wavelength intensity data from Optec Titan \citep{zhang2022introducing}. This methodology demonstrates the potential of transformer-based models to integrate a variety of point cloud features such as intensity, reflectance, amplitude, and echo deviation. However, it is important to note that research in this specific field is still limited, especially in the context of multispectral point clouds.

\subsection{Multispectral Point Cloud Data for Land Cover Mapping}

Early methods for land cover mapping with multispectral point clouds typically focused on utilizing rasterized data, i.e., Digital Surface Models (DSMs), Digital Terrain Models (DTMs), and intensity images, and hand-crafted features, such as various intensity ratios and indices \citep{karila2017feasibility,matikainen2017object,teo2017analysis}. These extracted features were subsequently fed into machine learning algorithms, commonly RF and SVM, or processed with rule-based algorithms. For example, \citet{matikainen2017object}  used Optech Titan data and combined multi-wavelength intensity images with features derived from DSM/DTM to map urban land cover using an object-based RF. Similarly, \citet{karila2017feasibility} utilized multispectral Optech Titan data with an RF classifier for road mapping. \citet{wichmann2015evaluating} used a point-based classification approach for multispectral point clouds that were obtained by merging data from three wavelengths of Optech Titan using a nearest neighbor approach. They also used a pseudo normalized difference vegetation index (NDVI). These previous studies showed the effectiveness of multispectral data for land cover mapping applications, and especially for distinguishing ground-level classes from each other.

Researchers have increasingly applied DL-based approaches to multispectral point cloud data too. A common approach involves rastering 3D point clouds into multi-layer 2D images, where each layer represents specific point cloud features such as intensity values from different wavelengths and height. For example, \citet{pan2020land} rasterized Optech Titan multispectral 3D LiDAR data into 2D images and used three-channel intensities and elevations as input to a CNN-based architecture for land cover classification in Ontario, Canada.

However, the rasterization process is inherently subject to information loss. To address this limitation, researchers have developed point cloud-specific networks to process 3D point cloud data directly.
\citet{jing2021multispectral} proposed an improved PointNet++ incorporating a Squeeze-and-Excitation (SE) block designed for multispectral LiDAR point cloud classification. \cite{zhao2021airborne} introduced a feature reasoning network using a graph convolution network (FR-GCNet) to classify airborne multispectral LiDAR point clouds from Optech Titan sensors. Furthermore,~\cite{zhang2022introducing} introduced an improved transformer to classify 3D multispectral point clouds directly.

Recent developments have focused on advanced architectures that more effectively utilize the multispectral characteristics of the data. \citet{xiao2022multispectral} modified RandLA-net for large-scale multispectral LiDAR point clouds of real-world land covers. \citet{li2022agfp} proposed a new type of convolution that combines attention mechanism and graph geometric moments convolution to extract local geometric features. In addition, they proposed  a feature up-sampling module to integrate features from different scales. \citet{yu2022capvit} developed CapViT that rasterizes multispectral point clouds into top-view feature images based on the point cloud attributes in different channels.

\section{Study Area and Data}
\subsection{Study area}
\begin{figure}[H]
    \centering
    \includegraphics[width=0.7\linewidth]{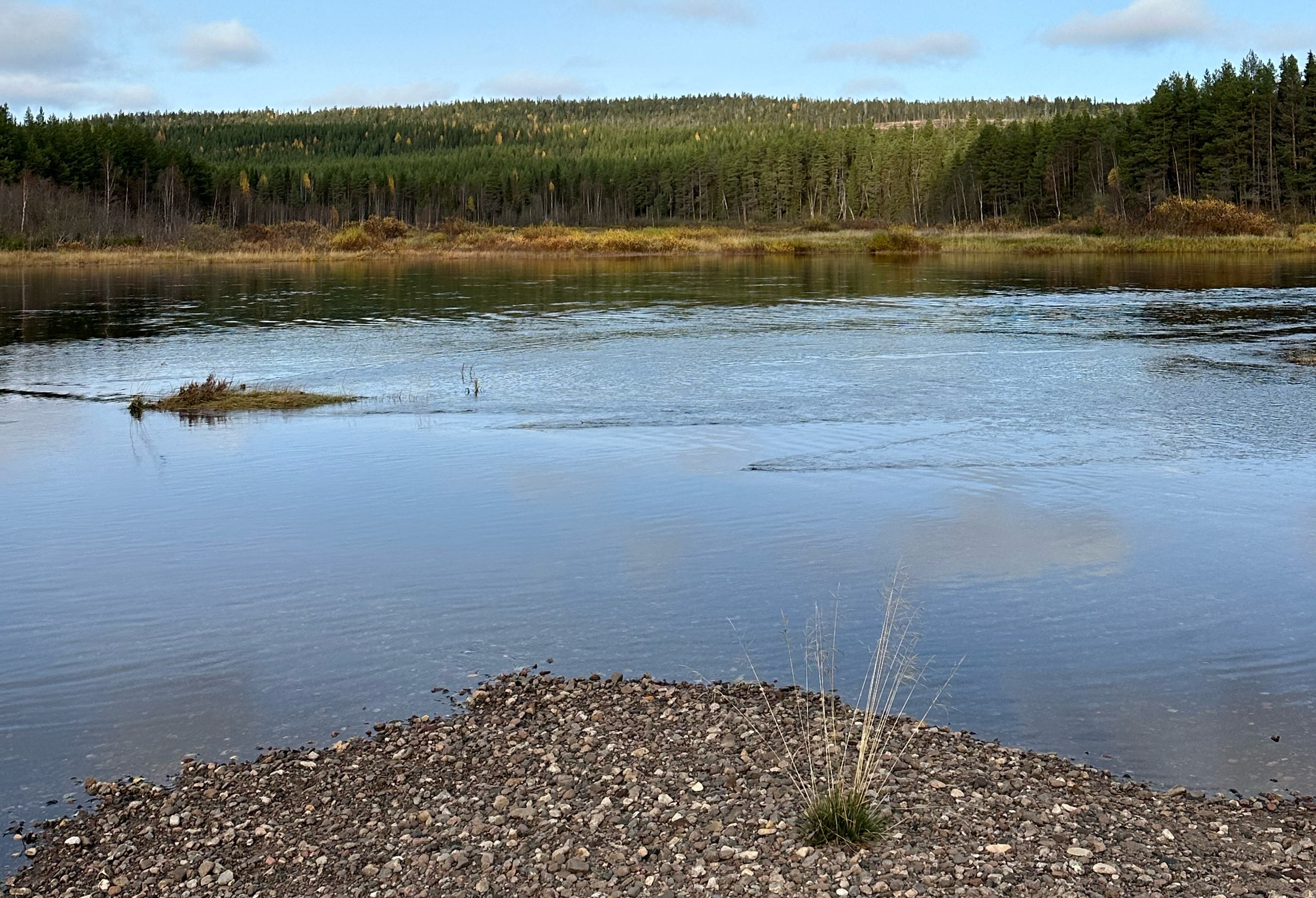}
    \caption{Oulanka river in the northeastern part of Finland }
    \label{fig: oulanka river}
\end{figure}

\begin{figure}[H]
    \centering
    \includegraphics[width=\linewidth]{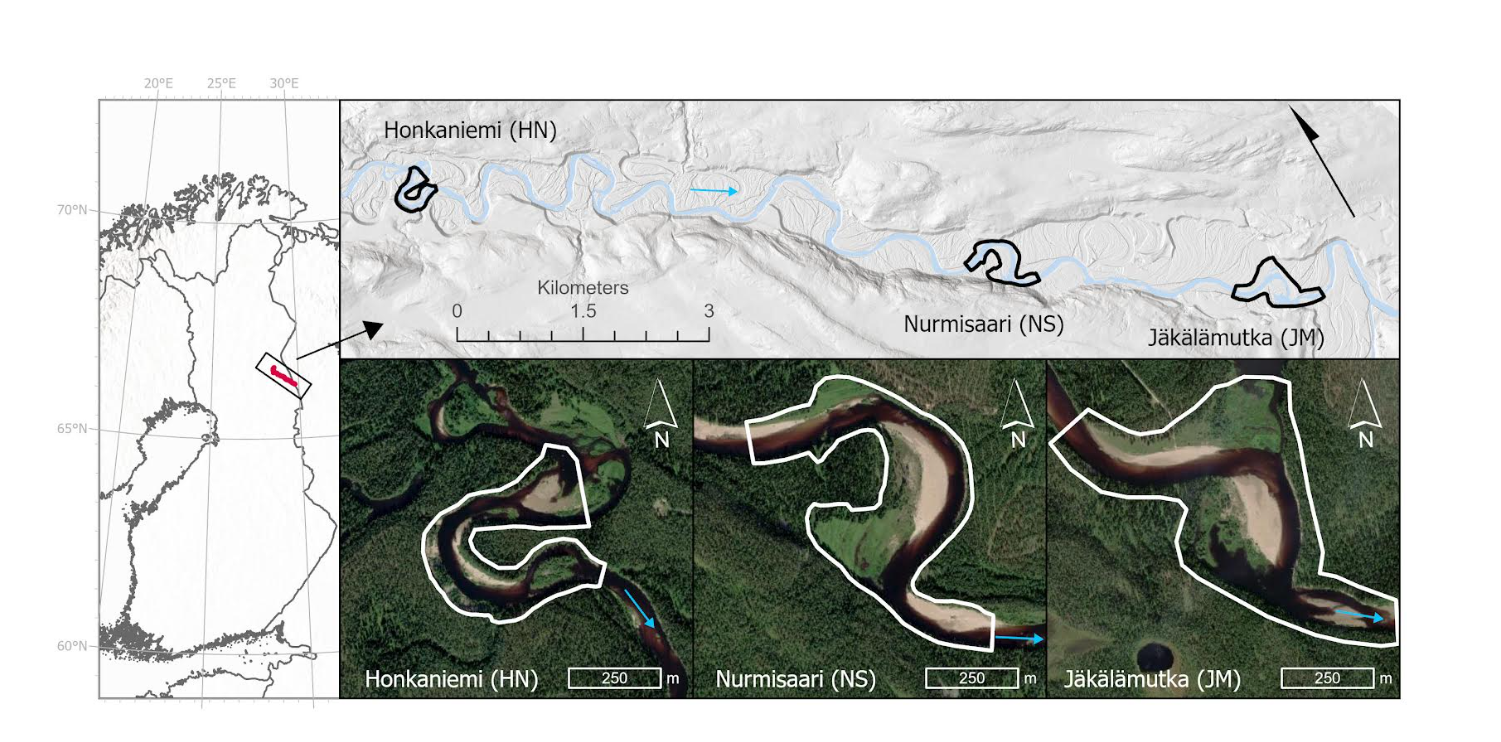}
    \caption{A map of Finland highlighting the three study areas: NS, HN, and JM.}
    \label{fig:test areas}
\end{figure}

The study area is located in the northeastern part of Finland at the Oulanka River (approximately 66.3 °N, 29.6 °E). Oulanka (Figure \ref{fig: oulanka river}) is a meandering river with sandy bedforms and regular flooding during spring when the snow melts, causing strong sediment erosion and deposition. The study focuses on three different river bends in the Oulanka River: Nurmisaari (NS), Honkaniemi (HN), and Jäkälämutka (JM). The study areas are shown in Figure \ref{fig:test areas}.

Each of the studied bends represents a distinct stage in meander evolution, primarily reflected in differences in sediment composition and point-bar morphology \citep{koutaniemi1979late,koutaniemi2000meanderointi}. Together, they illustrate a transition from a coarse, stable, and mature bar system to a finer, more dynamic, and flood-dominated environment within the Oulankajoki reach.

Nurmisaarenniemi, the oldest and most mature bend, is a compound asymmetric meander composed of coarse sediments, large pebbles, gravel, and gravelly sand. Its armoured point bar exhibits a clear downstream fining trend from pebbles at the bar head to sand toward the tail. The point bar surface and mid-channel bars are partly vegetated with low shrubs, while the surrounding floodplain hosts mature forest, reflecting long-term stability and reduced bar mobility.

Honkaniemi marks the most confined section of the river, forming a simple symmetric meander in gravel and coarse sand \citep{blaafield2024temporal}. The point bar is actively reworked by floods, showing alternating erosional and depositional zones and a subtle fining trend downstream. An armour layer of pebbles protects the bar head, whereas finer gravel and sand dominate the bar tail. Despite its active bar surface, the adjacent floodplain remains well vegetated and geomorphologically stable due to narrow valley confinement causing the point bar to migrate parallel to the valley axis.

Jäkälämutka, the youngest and most dynamic bend, is a simple asymmetric meander developed in fine to medium sand \citep{blaafield2024temporal}. Frequent inundation during seasonal floods produces diverse depositional microforms with spatially variable grain-size distributions. These features are continuously reshaped by hydraulic sorting and selective sediment transport due to varying flow dynamics, producing a gradual and heterogeneous textural pattern across the point-bar surface. The floodplain supports pioneer vegetation, reflecting continuous adjustment and rapid bar evolution.

\subsection{Multispectral LiDAR Data Acquisition}

\begin{figure}[H]
    \centering
    \includegraphics[width=0.5\linewidth]{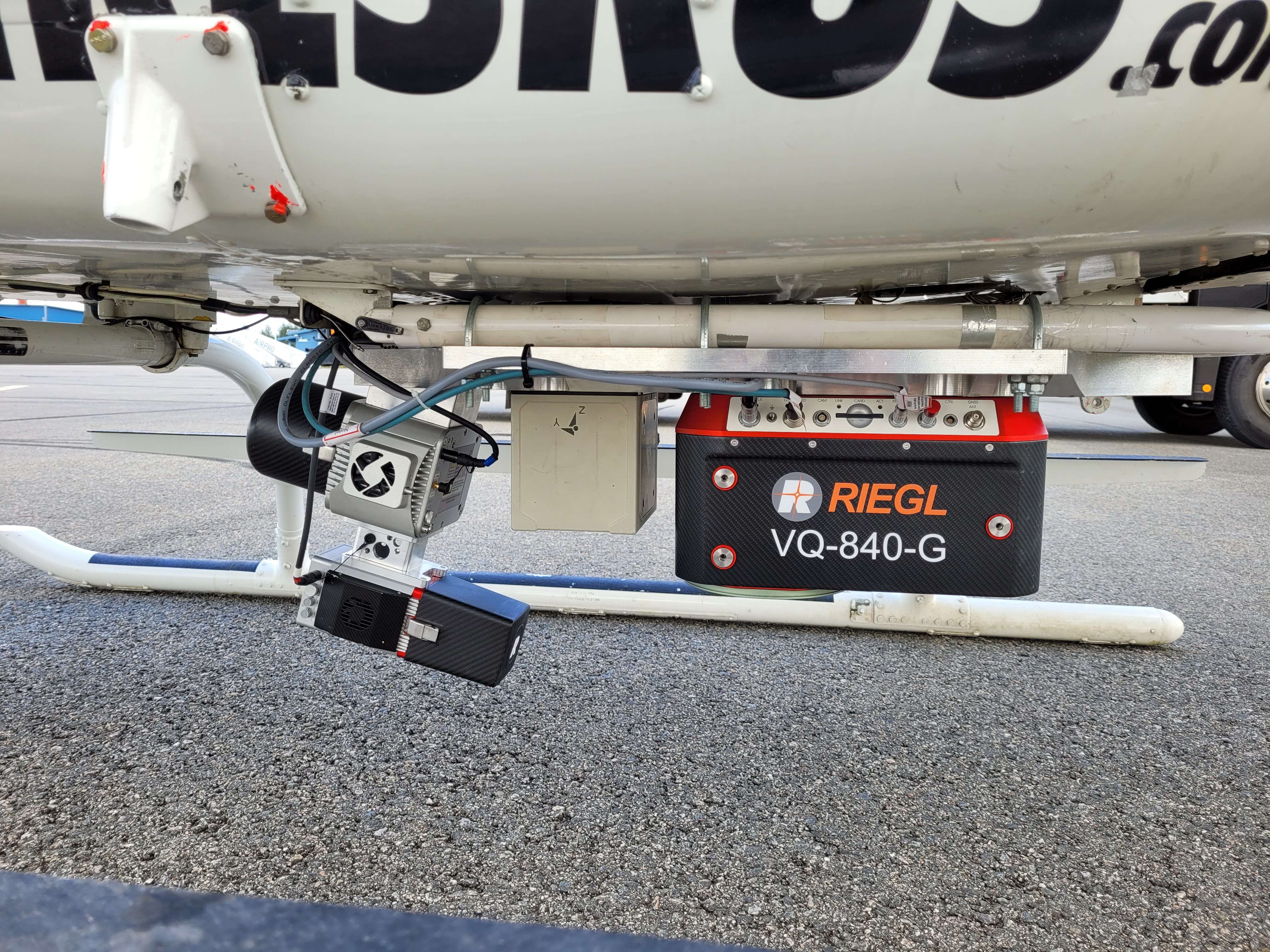}
    \caption{HeliALS-TW system}
    \label{fig:Heli}
\end{figure}

Multispectral laser scanning data was acquired on September 6th, 2022, using FGI's three-wavelength laser scanning system HeliALS-TW. The HeliALS-TW system as shown in Figure \ref{fig:Heli} consists of three separate Riegl (RIEGL Laser Measurement Systems GmbH, Austria) laser scanners: VUX-1HA (scanner 1), MiniVUX-3UAV (scanner 2), and VQ-840-G (scanner 3) operating at the wavelengths of 1550 nm, 905 nm, and 532 nm, with beam divergences of 0.5 mrad × 0.5 mrad, 0.5 mrad × 1.6 mrad, and 1.0 mrad × 1.0 mrad, respectively. HeliALS-TW was carried by a helicopter flying at approximately 100 m above the river at the speed of 15 m/s. The system's navigation data was used to compute the scanners' trajectories, which were used to compute the georeferenced point clouds produced by each scanner. The resulting raw point cloud data served as the input for subsequent preprocessing and analysis stages. It is worth noting that the point cloud density is very high compared to typical ALS data. 
Detailed specifications of each scanner in the HeliALS-TW system are shown in Table \ref{tab:scanner-specs}.

\begin{table}[H]
 \small
 \centering
 \begin{tabularx}{\textwidth}{@{}Xccc@{}}
  \toprule
  \textbf{Scanner} & \textbf{1} & \textbf{2} & \textbf{3} \\
  \textbf{Model} & VUX-1HA & miniVUX-3UAV& VQ-840-G \\
  \midrule
  Wavelength (nm) & 1,550 & 905 & 532 \\
  %Flight altitude above ground level (m) & 80/100& 80/100& 80/100\\
  Approx. point density (points/m$^2$) & 1,400& 500& 1,600\\
  Number of returns (max.) & 12& 5 & 5 \\
  Maximum scanning angle ($^\circ$) & 360 & 120& 28 $\times$ 40 \\
  Laser beam divergence (mrad) & 0.5 & 0.5 $\times$ 1.6 & 1\\
  Laser beam diameter at the ground level (cm) & 5 & 5 $\times$ 16 & 10\\
  Pulse repetition rate (kHz) & 1,017 & 300& 50\\
  \bottomrule
 \end{tabularx}
 \caption{The specifications of the laser scanners in the multispectral laser scanning system HeliALS for Oulanka river}
 \label{tab:scanner-specs}
\end{table}

\subsection{Data Preprocessing}
Data preprocessing involves multiple sequential steps: downsampling, noise filtering, ground segmentation, multispectral point cloud creation, data annotation, and normalization.

\paragraph{\textbf{Downsampling}}
The raw point cloud data were downsampled using a voxel grid with a resolution of 2 cm to remove redundant points and create evenly distributed points. The 2 cm resolution was selected as a compromise between preserving fine-scale riverine cover such as sand and gravel and increasing computational efficiency. The downsampling was performed using Laspy and Numpy libraries and the Python programming language.

\paragraph{\textbf{Noise Filtering}}
Even though LiDAR scanners usually have high accuracy and precision, they can still introduce noise. These outliers can impact subsequent data processing and analyses. To address this problem, we used the Statistical Outlier Removal (SOR) \citep{pdal_contributors_2022_2616780} method.
The SOR computes the mean distance of each point to its k nearest neighbors. After establishing a global mean distance and standard deviation, the threshold interval is computed with the assumption that the distribution of point-to-neighbor distances is characterized by a Gaussian distribution. Any points with the mean distances outside the threshold interval are considered outliers and thus are removed. In this study, k and the deviation multiplier were both set to 10.

\paragraph{\textbf{Ground Segmentation}}
We applied the Cloth Simulation Filter (CSF) \citep{zhang2016easy} algorithm provided by PDAL to segment ground points. The CSF algorithm classifies ground points by simulating a flexible cloth draped over an inverted terrain surface. The CSF was configured with a cloth resolution of 1, a distance threshold of 1.5, a time step of 0.65, and a rigidness of 1.

\paragraph{\textbf{Creating Multispectral Point Clouds}}
To create comprehensive multispectral cloud datasets, we integrated data from the three different wavelengths. Our approach uses a nearest-neighbor technique using K-Dimensional (KD) trees \citep{bentley1975multidimensional} to combine data from different scanners.
For each point in each scanner, we identify neighboring points from the other scanners within a 0.25-meter radius. The resulting feature vector for each point includes the XYZ coordinates, three-channel intensities, three-channel reflectances, three-channel amplitudes, three-channel deviations, the number of returns, and return numbers. Utilizing the KD-tree, we address the varying point densities from different scanners. In areas where one scanner has limited data, information can be supplemented from denser measurements of other scanners. This method ensures that every point in the final multispectral point clouds has a feature vector as complete as possible. This output dataset provides high-quality, georeferenced multispectral point clouds with geometric and spectral information, suitable for detailed mapping of riverine land covers.

\subsection{Point Cloud Attribute}
After processing, the complete point cloud has attributes/features from combining three different wavelengths. Each point cloud contains 3D coordinates, intensities, reflectances, amplitudes, deviations from three channels, return number, and the number of returns. Each attribute is described in Table \ref{tab:point_cloud_attributes}. Further background on LiDAR features can be found in \cite{wehr1999airborne} and \cite{harding2001laser}. Reflectance, deviation, and amplitude calculations can be found in \citet{pfennigbauer2010improving}.

\begin{table}[H]
  \centering
  \small
  \begin{tabularx}{\textwidth}{@{}lX@{}} 
    \toprule
    \textbf{Point Attribute} & \textbf{Description} \\
    \midrule
    3D coordinates & X, Y, Z positional data from LiDAR scanning \\
    Intensity & Strength of the return signal \\
    Reflectance & Surface reflectivity \\
    Amplitude & Peak amplitude of the return waveform \\
    Deviation & The deviation of the return echo waveform from the system impulse response  \\
    Return number & Sequential identifier of each return from one laser pulse \\
    Number of returns & Total count of returns detected per each illumination pulse \\
    \bottomrule                   
  \end{tabularx}
  \caption{Point cloud attributes}
  \label{tab:point_cloud_attributes}
\end{table}

\subsection{Data Annotation}
\begin{figure}[H]
    \centering
    \includegraphics[width=1\linewidth]{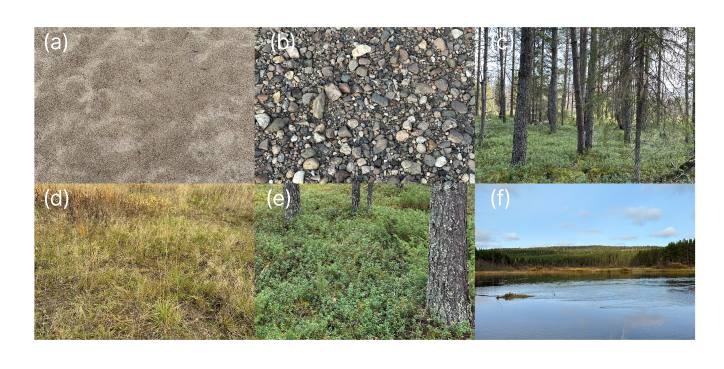}
    \caption{Example images of the six land cover classes used in the dataset (a) sand, (b) gravel, (c) high vegetation, (d) low vegetation, (e) forest floor, (f) water}
    \label{fig:gt-pics}
\end{figure}

Data was annotated into six different classes: sand, gravel, low vegetation, high vegetation, forest floor, and water. Data annotation was done based on visual identification on the high-resolution orthophoto using QGIS (more details of orthophotos in section 3.6), except for the forest floor class, where the annotated points were first obtained with ground segmentation and later filtered to exclude points outside of high vegetation zones. HN dataset was annotated using a sparse labeling approach, where only selected regions were manually labeled to reduce annotation costs while maintaining class diversity. More details about each land cover class are provided in Table~\ref{tab:landcover}, and visual examples can be found in Figure~\ref{fig:gt-pics}.

\begin{table}[H]
\centering
\small
\begin{tabular}{ll}
\hline
\textbf{Class} & \textbf{Description} \\
\hline
Sand & Loose, granular surface material \\
Gravel & Coarse rock fragments \\
High vegetation & Tall, dense vegetation such as trees \\
Low vegetation & Shorter vegetation, grasses, or ground cover \\
Forest floor & Moss or other low vegetation under high vegetation\\
Water & Rivers and water bodies \\
\hline
\end{tabular}
\caption{Land cover classes and their characteristics.}
\label{tab:landcover}
\end{table}

\subsection{Dataset description}
Three datasets from different riverine environments were used in this study. The details of the datasets are shown in Table \ref{table:characteristic}, and the class distributions are presented in  Table \ref{table:datadist}. Each dataset is described in detail below.

\begin{figure}[H]
    \centering
    % Row 1
    \begin{subfigure}[b]{0.45\textwidth}
        \centering
        \includegraphics[width=0.7\textwidth]{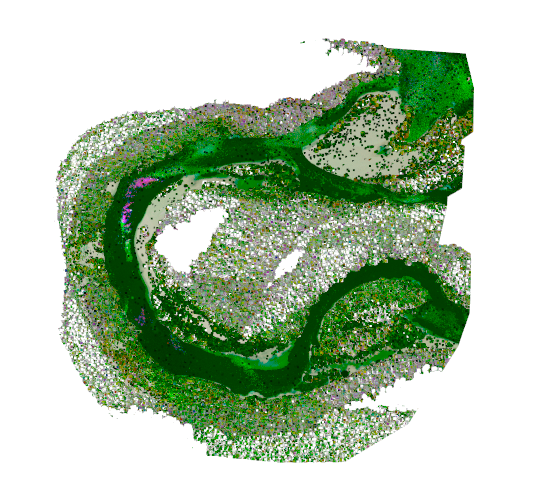}
        \caption{multispectral NS point clouds}
        \label{fig:subNS1}
    \end{subfigure}
     %\hfill
    \begin{subfigure}[b]{0.4\textwidth}
        \centering
        \includegraphics[width=0.7\textwidth]{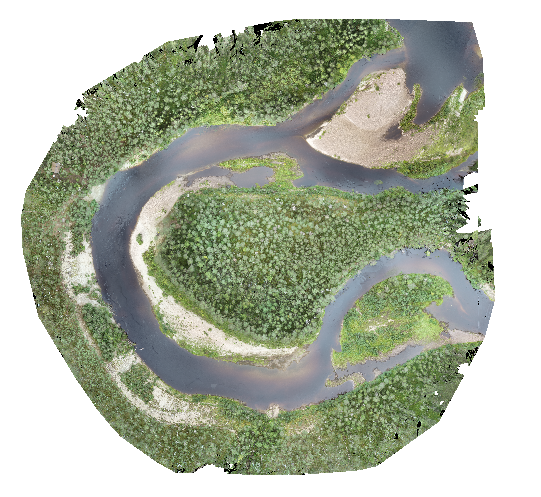}
        \caption{NS orthophoto}
        \label{fig:subNS2}
    \end{subfigure}
    
    %\vspace{0.5cm} % Vertical space between rows  
    % Row 2
    \begin{subfigure}[b]{0.4\textwidth}
        \centering
        \includegraphics[width=0.7\textwidth]{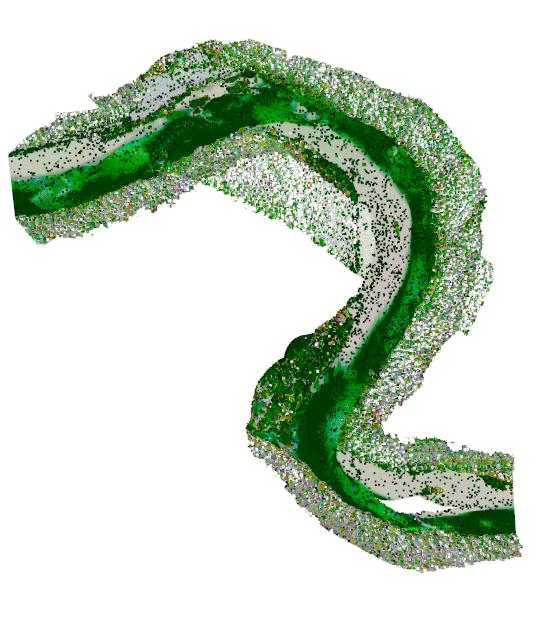}
        \caption{multispectral HN point clouds}
        \label{fig:subHN3}
    \end{subfigure}
     %\hfill
    \begin{subfigure}[b]{0.4\textwidth}
        \centering
        \includegraphics[width=0.7\textwidth]{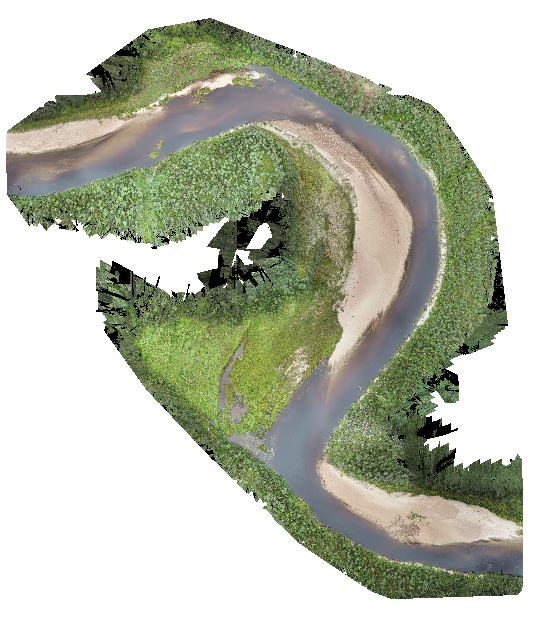}
        \caption{HN orthophoto}
        \label{fig:subHN4}
    \end{subfigure}
    %\vspace{0.5cm} % Vertical space between rows
    % Row 3
    \begin{subfigure}[b]{0.48\textwidth}
        \centering
        \includegraphics[width=0.8\textwidth]{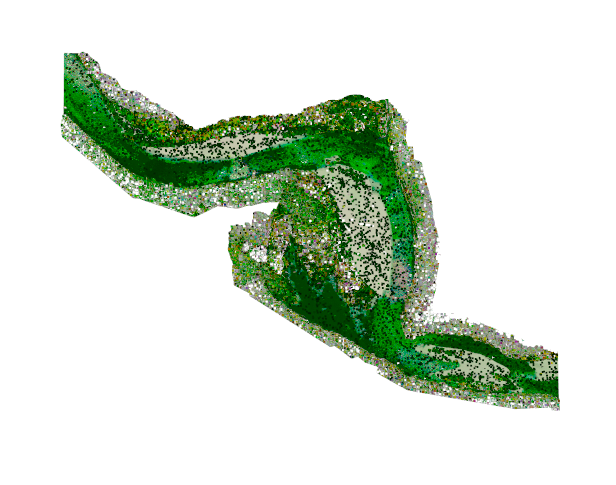}
        \caption{multispectral JM point clouds}
        \label{fig:subJM5}
    \end{subfigure}
     %\hfill
    \begin{subfigure}[b]{0.38\textwidth}
        \centering
        \includegraphics[width=0.8\textwidth]{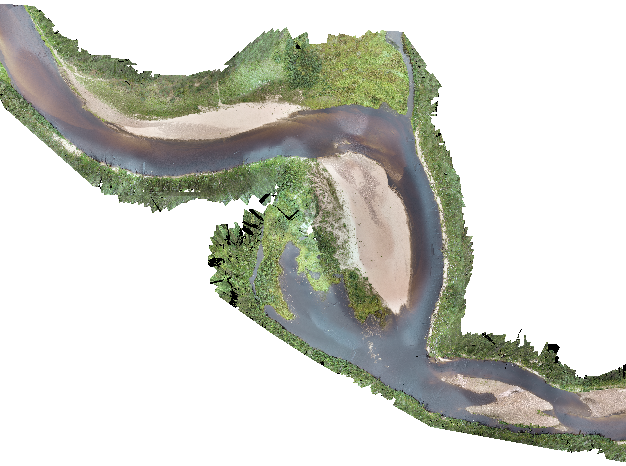}
        \caption{JM orthophoto}
        \label{fig:subJM6}
    \end{subfigure}
    \caption{Oulanka point cloud data colored by 3-channel intensities as RGB and the corresponding orthophotos}
    \label{fig:data-all}
\end{figure}

\paragraph{\textbf{Dataset Nurmisaari (NS)}}
This dataset contains about 34 million points. The sand class is only about 0.5\% making this class underrepresented in this area. The NS point clouds are illustrated in Figure \ref{fig:subNS1} and the corresponding orthophoto is shown in Figure \ref{fig:subNS2}.

\paragraph{\textbf{Dataset Honkaniemi (HN)}}
This dataset is smaller, containing approximately 5.2 million points for training and about 1.3 million points for validation. Unlike the NS dataset, the sand class is well represented with about 32\% of the sand class. The HN point clouds are illustrated in Figure \ref{fig:subHN3} and the corresponding orthophoto is shown in Figure \ref{fig:subHN4}. It is worth noting that about 7.5\% of all points in the HN area are included in this dataset.

\paragraph{\textbf{Dataset Jäkälämutka (JM)}}
This dataset is used exclusively for testing and evaluating the generalization ability of the trained models. It contains approximately 10 million annotated points with a relatively balanced distribution across classes. The JM point clouds are illustrated in Figure \ref{fig:subJM5} and the corresponding orthophoto is shown in Figure \ref{fig:subJM6}.

\begin{table}[H]
\small
  \centering
  \begin{tabular}{@{}lrrr@{}}   % l = left-align label; rrr = right-align numbers
    \toprule
    \diagbox{Class}{Site} & Nurmisaari (NS) & Honkaniemi (HN) & Jäkälämutka (JM) \\
    \midrule            % ← uncomment if you want a rule below the header
    Sand         &  0.5 & 32.3 & 30.2 \\
    Gravel       & 10.1 &  7.7 & 11.8 \\
    High Vegetation & 59.6 & 35.2 & 25.8 \\
    Low Vegetation &  5.3 &  3.3 & 13.0 \\
    Forest Floor & 11.6 &  5.2 &  9.4 \\
    Water        & 12.9 & 16.2 &  9.8 \\
    \bottomrule
  \end{tabular}
  \caption{Class distribution (\%) of annotated data in the three study areas}
  \label{table:datadist}
\end{table}

The model is trained on a combination of the NS and HN datasets: NS has a large amount of points but contains a very low proportion of annotated sediment classes, while HN complements it with richer sediment annotations, as shown in Table \ref{table:datadist}.

\begin{table}[H]
  \centering
  \small
  \begin{tabularx}{\textwidth}{@{}lXXX@{}} 
    \toprule
    \diagbox{Detail}{Site} & Nurmisaari (NS) & Honkaniemi (HN) & Jäkälämutka (JM) \\
    \midrule        
    East-west (m)        &  498 & 672  & 919  \\
    North-south (m)       & 477 & 982  & 680 \\
    Area (m2)     & 238000  & 525000  & 625000  \\
    Number of points (millions)     & 34   & 6.5   & 10   \\
    Usage      &  Train &  Train/Validation & Test \\
    \bottomrule
  \end{tabularx}
  \caption{Details of the three study areas}
  \label{table:characteristic}
\end{table}

Orthophotos shown in Figures \ref{fig:subNS2}, \ref{fig:subHN4}, and \ref{fig:subJM6} were generated and processed by the Fluvial and Coastal Research Group at the Department of Geography and Geology, University of Turku, with Pix4Dmapper using Structure-from-Motion protocol detailed in \citet{Micheletti2015}. The resolutions of the orthophotos: NS is 1.92 cm, HN is 2.08 cm, and JM is 1.92 cm.

\section{Methodology}

This section outlines our approach for automated land cover mapping through semantic segmentation of multispectral LiDAR point clouds using PTv2. This section describes the multi-dataset training framework and the evaluation metrics.

\subsection{Multi-Dataset Training framework}
Supervised learning in remote sensing often faces two critical challenges. One is limited availability of high-quality annotated data, and the other one is poor generalization to new environments because of domain shift. In river environments, those domain shifts can arise from sediment compositions or variations in terrain morphology. In our study, all datasets are collected from the Oulanka River, which covers different river bends at different stages in meander evolution. These diverse sediment compositions and point-bar morphologies result in distributional differences between datasets, naturally creating domain shifts and hindering model generalization.

To address these problems while not relying on laborious extensive data annotation, we utilized multi-dataset training, specifically by training the model on the NS dataset, which is large but contains a low proportion of annotated data for sediment classes, and the sparsely annotated HN dataset. Despite covering only limited areas, HN contains a higher proportion of annotated data for rare classes, particularly sand and gravel. Additionally, to mitigate class imbalance, we applied a weighted cross-entropy loss function, assigning higher penalties to misclassified data points from underrepresented classes.

During training, the model was trained on both datasets simultaneously. We constructed each mini-batch to contain data from both datasets. This approach allows the model to learn from a more diverse set of examples of critical but infrequent classes such as sand and gravel. By training the model on river land covers from different geographical areas, we aim to improve its ability to generalize to unseen data such as our test set (Dataset JM).

\subsection{Experimental Settings}
The deep learning models are implemented using PyTorch and trained on 2 NVIDIA A100 GPUs. We use Adaptive Moment Estimation with Decoupled Weight Decay Regularization (AdamW) \citep{loshchilov2017decoupled} with a learning rate of 0.001 and a weight decay of 0.075, using a cosine annealing schedule. The models are trained for 50 epochs using a batch size of 32. To improve model robustness, we apply data augmentation techniques, including random scaling and translation. Augmentation techniques are summarized in Table \ref{tab:data_augmentation}.

\begin{table}[H]
    \centering
    \begin{tabular}{@{}lll@{}}
        \toprule
        \textbf{Augmentation} & \textbf{Range} & \textbf{Description} \\
        \midrule
        Scaling      &  0.8--1.2 & Random uniform resizing of the point cloud \\
        Translation  & $\pm$0.2 units & Shifts all points along x, y, and z axes \\
        \bottomrule
    \end{tabular}
    \caption{Data augmentation techniques applied during training.}
    \label{tab:data_augmentation}
\end{table}

\subsection{Evaluation Metrics}
To assess the performance of PTv2 for land cover mapping based on semantic segmentation, we chose evaluation metrics commonly used for semantic segmentation tasks: Intersection over Union (IoU), accuracy, and precision. The combination of IoU, accuracy, and precision provides a comprehensive evaluation of both spatial and point-level classification performance.

IoU is the intersection of predicted and ground truth points divided by their union. IoU quantifies how well the model's prediction aligns with the actual regions of different land cover types in the point cloud data. For a given land cover class $c$, IoU is calculated as
\begin{equation}
    \text{IoU}_c = \frac{\text{TP}_c}{\text{TP}_c + \text{FP}_c + \text{FN}_c},
\end{equation}
where $\text{TP}_c$, $\text{FP}_c$, and $\text{FN}_c$ denote true positives, false positives, and false negatives of points belonging to class  class $c$.

The IoU metric ranges from 0 to 1, where 0 indicates no overlap between prediction and ground truth, and 1 indicates perfect overlap. The mean IoU (mIoU) across all classes can provide an overall performance of segmentation quality, and it can be calculated as :
\begin{equation}
    \text{mIoU} = \frac{1}{N} \sum_{c=1}^{N} \text{IoU}_c
\end{equation}
where $N$ is the total number of land cover classes.

Accuracy represents the proportion of correctly classified points. For each class $c$, the accuracy is calculated as:
\begin{equation}
    \text{Accuracy}_c = \frac{\text{TP}_c}{\text{TP}_c + \text{FN}_c}
\end{equation}
This is also equivalent to recall. In addition, we also calculate mean accuracy as the average of per-class accuracies. The mean accuracy across all classes is computed as:
\begin{equation}
    \text{Mean Accuracy} = \frac{1}{N} \sum_{c=1}^{N} \text{Accuracy}_c
\end{equation}

Precision measures the model's ability to correctly identify points of a specific class without false positives. For each class $c$, precision is calculated as:
\begin{equation}
    \text{Precision}_c = \frac{\text{TP}_c}{\text{TP}_c + \text{FP}_c}
\end{equation}
This metric represents the proportion of points predicted as class $c$ that actually belong to that class.

\section{Results and Discussion}

This section presents the experimental evaluation of PTv2 for land cover mapping with semantic segmentation in riverine environments using multispectral LiDAR point clouds. We evaluated model holistic performance using mean accuracy (mAcc), mean IoU, and mean precision for all six riverine land cover classes. Due to the class imbalance present in the data, we also reported class-wise accuracy, class-wise IoU, and class-wise precision to analyze performance for each individual land cover type.

We conducted ablation studies to investigate the contribution of different LiDAR spectral wavelengths and LiDAR features and to identify the optimal combination of input features. In addition, we evaluated the effectiveness of the multi-dataset training with sparse annotations on the model's ability to generalize, with the objective of enhancing robustness and adaptability to diverse river environments.

The following subsections below describe our experimental settings, multispectral feature analysis, feature combinations, evaluation of multi-dataset training, comparative results to other established machine learning baselines, and visualizations of predictions on the test data. All numerical evaluations were performed on the JM test dataset.

\subsection{Importance of Multispectral Data}

Understanding the contribution of different LiDAR channel combinations to semantic segmentation performance on various land cover classes is crucial. Thus, we conducted ablation experiments using five channel sets, as shown in Table ~\ref{tab:channel_features}, starting from the geometric coordinates baseline (XYZ) and incrementally adding spectral channels (1-channel, 2-channel, and 3-channel spectral features). The full configuration, all features, has all available features, including return number and number of returns. Table ~\ref{tab:channel_IOU}, Table~\ref{tab:channel_ACC}, and Table~\ref{tab:channel_precision} present the quantitative results using IoU, accuracy, and precision metrics, respectively.

\begin{table}[H]
    \centering
    \resizebox{0.95\textwidth}{!}{%
    \begin{tabular}{p{3cm}p{12cm}}
        \toprule
        \textbf{Channel Set} & \textbf{Feature configuration} \\
        \midrule
        XYZ & • 3D coordinates \\
        \addlinespace[0.3em]
        Channel 1 & • 3D coordinates \newline • 1-channel intensity, reflectance, amplitude, deviation (all from VUX) \\
        \addlinespace[0.3em]
        Channels 1--2 & • 3D coordinates \newline • 2-channel intensity, reflectance, amplitude, deviation (all from VUX and miniVUX) \\
        \addlinespace[0.3em]
        Channels 1--3 & • 3D coordinates \newline • 3-channel intensity, reflectance, amplitude, deviation (all from VUX, miniVUX, VQ) \\
        \addlinespace[0.3em]
        All features & • 3D coordinates \newline • 3-channel intensity, reflectance, amplitude, deviation (all from VUX, miniVUX, VQ) \newline • Return number, number of returns \\
        \bottomrule
    \end{tabular}
    }
    \caption{Feature configurations for different channel sets in the multispectral point cloud ablation study}
    \label{tab:channel_features}
\end{table}

\begin{table}[H]          
  \centering
  \small
  \setlength{\tabcolsep}{4pt}
  \renewcommand{\arraystretch}{1.1}
  \begin{tabular}{@{}l c c c c c c c@{}} 
    \toprule
    {Channel set}  & {\textbf{mIoU}} & {Sand} & {Gravel} & {High veg.} & {Low veg.}
                    & {Forest floor} & {Water} \\
    \midrule      
    XYZ            & 0.643 & 0.487 & 0.194 & 0.874 & 0.685 & 0.765 & 0.852 \\
    Channel~1      & 0.865 & 0.849 & 0.908 & 0.905 & 0.773 & 0.763 & 0.993 \\
    Channels~1–2   & 0.937 & 0.988 & 0.991 & 0.903 & 0.934 & 0.813 & 0.993 \\
    Channels~1–3   & 0.938 & 0.993 & 0.971 & 0.910 & 0.938 & 0.818 & 0.999 \\
    All features   & 0.950 & 0.990 & 0.971 & 0.931 & 0.930 & 0.878 & 0.998 \\
    \bottomrule
  \end{tabular}
  \caption{Ablation study results showing \textbf{IoU} for each class and \textbf{mIoU} across all classes on the JM test dataset}
  \label{tab:channel_IOU}
\end{table}

\begin{table}[H]   
  \centering
  \small
  \setlength{\tabcolsep}{4pt}
  \renewcommand{\arraystretch}{1.1}
  \begin{tabular}{@{}l c c c c c c c@{}}
      \toprule
      {Channel set} & {\textbf{mAcc}} & {Sand} & {Gravel} & {High veg.} &
      {Low veg.} & {Forest floor} & {Water} \\

      \midrule     
      XYZ                  & 0.751 & 0.733 & 0.258 & 0.893 & 0.818 & 0.857 & 0.949 \\
      Channel~1            & 0.932 & 0.849 & 0.984 & 0.924 & 0.912 & 0.924 & 0.998 \\
      Channels~1–2         & 0.970 & 0.988 & $\approx 1.000$ & 0.914 & 0.972 & 0.950 & 0.999 \\
      Channels~1–3         & 0.971 & 0.994 & $\approx 1.000$ & 0.919 & 0.968 & 0.948 & $\approx 1.000$ \\
      All features         & 0.977 & 0.990 & 0.998 & 0.938 & 0.970 & 0.966 & 0.999 \\
      \bottomrule
  \end{tabular}
  \caption{Ablation study results showing \textbf{Acc} for each class and \textbf{mAcc} across all classes on the JM test dataset}
  \label{tab:channel_ACC}
\end{table}

The baseline configuration using only 3D coordinates achieved modest performance (mIoU: 0.643, mAcc: 0.751). When including Channel 1 resulted in significant improvement in the performance (mIoU: 0.865, mAcc: 0.932), indicating an increase of about 0.222 mIoU unit and 0.181 mAcc unit, respectively. Incorporating both Channel 1 and Channel 2 further improved the performance (mIoU: 0.937, mAcc: 0.970). When combining Channel 1, Channel 2, and Channel 3, we observed only marginal gain in the result (mIoU: 0.938, mAcc: 0.971). The full feature configuration has resulted in the best semantic segmentation performance with an mIoU of 0.950 and an mAcc of 0.977.

Class-wise analysis shows that sand and gravel benefit the most from the spectral information. For example, the sand class improved from an IoU of 0.487 and Acc of 0.733 under the baseline configuration to an IoU of 0.990 and Acc of 0.990 under the full configuration. Likewise for the gravel class, the gravel class improved from an IoU of 0.194 and an Acc of 0.258 under the baseline to an IoU of 0.971 and an mAcc of 0.998 with the full configuration.

\begin{figure}[H]
    \centering
    \includegraphics[width=0.85\linewidth]{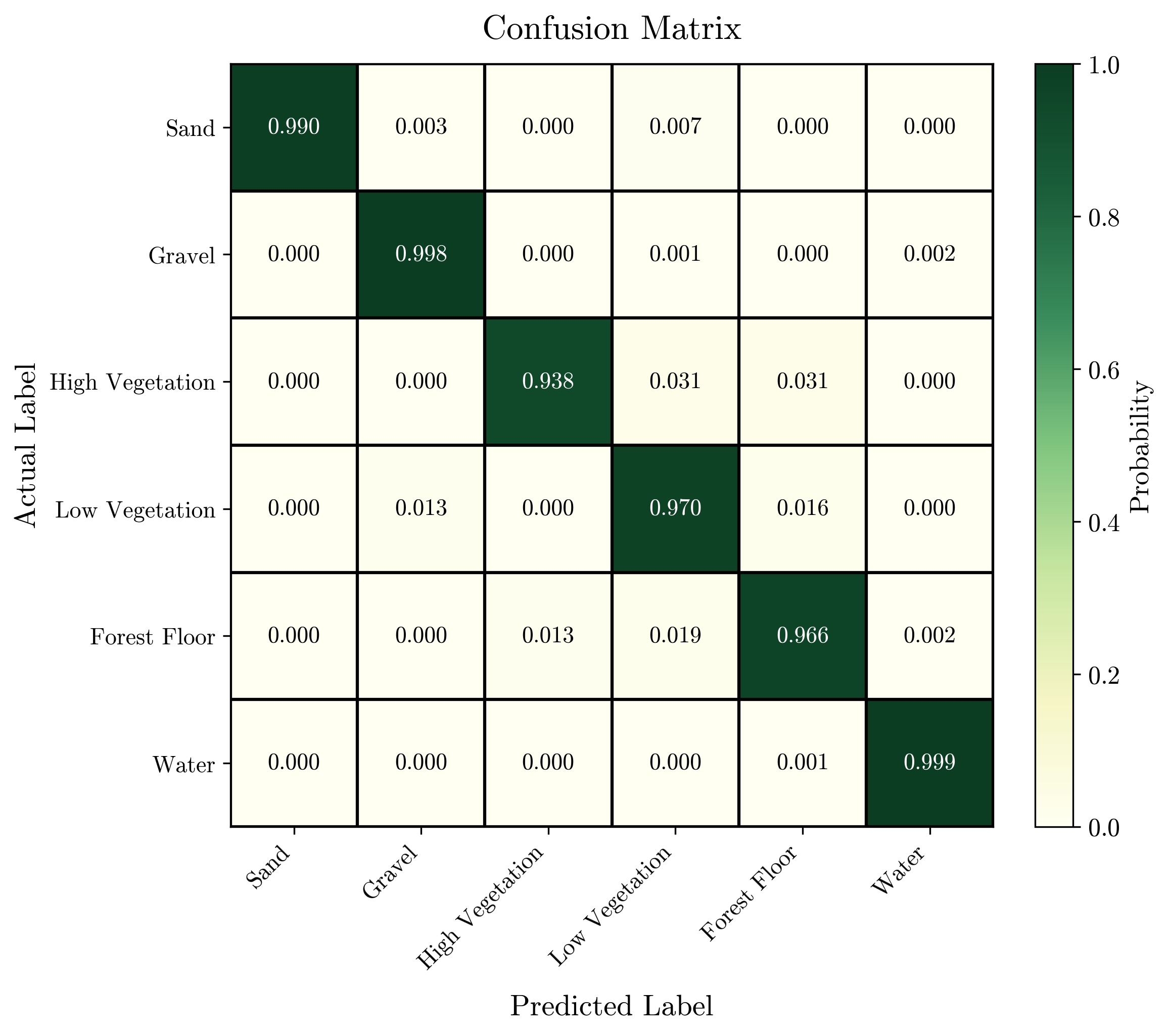}
    \caption{Confusion matrix for using all spectral features together with pulse information (row-wise normalization). Values shown are rounded to three decimal places.}
    \label{fig:cm_full}
\end{figure}

Water segmentation performed relatively well in all configurations, indicating that the water class can be robustly segmented even with limited spectral information. Vegetation classes also performed well in all configurations after the geometric baseline, with a marginal increase from additional features. Forest floor class performance showed moderate improvement with richer feature configurations, indicating a positive impact of spectral and pulse features. The confusion matrix for the model using the full feature configuration is illustrated in Figure \ref{fig:cm_full}. The matrix was normalized row-wise, where each element represents the proportion of points from an actual class that were assigned to each predicted class.

\begin{table}[H]         
  \centering
  \small
  \setlength{\tabcolsep}{2pt}
  \renewcommand{\arraystretch}{1.0}
  \begin{tabular}{@{}l c c c c c c c@{}}  
      \toprule
      {Channel set} & {\textbf{mPrecision}} & {Sand} & {Gravel} & {High veg.} & {Low veg.}
                    & {Forest floor} & {Water} \\
      \midrule      
      XYZ                     & 0.764 & 0.592 & 0.438 & 0.976 & 0.808 & 0.877 & 0.892 \\
      Channel~1               & 0.924 & $\approx 1.000$ & 0.922 & 0.978 & 0.835 & 0.814 & 0.995 \\
      Channels~1–2            & 0.964 & $\approx 1.000$ & 0.992 & 0.986 & 0.960 & 0.850 & 0.994 \\
      Channels~1–3            & 0.963 & $\approx 1.000$ & 0.971 & 0.985 & 0.969 & 0.857 & 0.999 \\
      All features            & 0.971 & $\approx 1.000$ & 0.973 & 0.992 & 0.958 & 0.906 & 0.999 \\
      \bottomrule
  \end{tabular}
  \caption{Ablation study results showing \textbf{Precision} for each class and \textbf{mean Precision} across all classes on the JM test dataset}
  \label{tab:channel_precision}
\end{table}

Precision scores (Table~\ref{tab:channel_precision}) provide additional validation to these findings, with the full feature set achieving the highest mean precision of 0.971. The additional spectral information has eliminated false positives for the sand class, which achieves perfect precision (up to roundoff error) across all configurations with spectral information. In addition, the precision of the gravel class increases significantly from 0.438 to 0.973, which further indicates that spectral information is useful for sediment mapping, leading to a minimum number of false positives. 

\subsection{Contribution of Spectral Features}
Understanding the contribution of each spectral feature to semantic segmentation performance is essential. Thus, we conducted an ablation study by training models with each spectral feature (from all 3 scanners) separately.

\begin{table}[H]      % use * for full-width in a two-column layout
  \centering
  \setlength{\tabcolsep}{4pt}
  \small
  \begin{tabular}{@{}l c c c c c c c@{}}
      \toprule
      {Feature} & {\textbf{mIoU}} & {Sand} & {Gravel} & {High veg.} &
      {Low veg.} & {Forest floor} & {Water} \\

      \midrule      
      Intensity     & 0.937 & 0.995 & 0.980 & 0.909 & 0.940 & 0.802 & 0.999 \\
      Reflectance   & 0.934 & 0.993 & 0.978 & 0.894 & 0.930 & 0.813 & 0.998 \\
      Amplitude     & 0.767 & 0.605 & 0.400 & 0.908 & 0.874 & 0.817 & 0.999 \\
      Deviation     & 0.719 & 0.606 & 0.222 & 0.883 & 0.877 & 0.727 & 0.997 \\
      \bottomrule
  \end{tabular}
  \caption{Ablation study results showing \textbf{IoU} and \textbf{mIoU} for each spectral feature. Each model variant is trained using only one input feature at a time with geometry to evaluate the individual contribution of each feature to semantic segmentation performance.}
  \label{tab:feature_ablation}
\end{table}

Table~\ref{tab:feature_ablation} presents the class-wise IoU scores and the mIoU scores for each spectral feature. The use of intensity and reflectance demonstrated the highest mIoU of 0.937 and 0.934, respectively. Both features showed strong performance for all the riverine land cover classes, especially for sand (IoU $> 0.99$) and gravel (IoU $\approx 0.98$) segmentation. This suggests that intensity and reflectance features are essential for differentiating riverine sediment classes.

In contrast, the use of amplitude and deviation resulted in significantly lower performance, with mIoU of 0.767 and 0.719, respectively. The performance degradation can be observed in sediment classes. Using amplitude features only resulted in an IoU of only 0.605 and 0.400 for sand and gravel, respectively. Similarly, deviation features resulted in  mIoU of 0.606 and 0.222 for the same classes. Nevertheless, vegetation and  water segmentation remained consistently strong across all spectral features, with mIoU ranging from 0.883 to 0.909 for the high vegetation class, mIoU ranging from 0.874 to 0.940 for the low vegetation class, and mIoU ranging from 0.997 to 0.999 for the water class. These findings suggest that intensity and reflectance are the most critical features for accurate land cover mapping, while amplitude and deviation may only provide limited benefits when it comes to riverine sediment segmentation classes.

\subsection{Impact of Multi-Dataset Training}
To evaluate the model's ability to generalize to diverse riverine environments under limited data availability, we assessed the impact of multi-dataset training by comparing models trained only on the NS dataset with those trained on NS and HN data (the default setting used in all other experiments). All models were trained using the full feature set and evaluated on the JM test set, which represents an unseen domain.

\begin{table}[H] 
  \centering
  \small
  \begin{tabular}{@{}l c c c c c c c@{}}
      \toprule
      {Training data} & {\textbf{mIoU}} & {Sand} & {Gravel} & {High veg.} &
      {Low veg.} & {Forest floor} & {Water} \\
      \midrule     
      NS        & 0.773 & 0.857 & 0.836 & 0.895 & 0.571 & 0.486 & 0.994 \\
      NS + HN   & 0.950 & 0.990 & 0.971 & 0.931 & 0.930 & 0.878 & 0.998 \\
      \bottomrule
  \end{tabular}
  \caption{Per-class \textbf{IoU} and \textbf{mIOU} performance comparison comparing models trained on the NS dataset alone and combined NS+HN dataset}
  \label{tab:multi_dataset_training}
\end{table}

Table~\ref{tab:multi_dataset_training} presents IoU scores of riverine land cover mapping between using a single dataset (NS) and a multi-dataset (NS and HN). The model trained only on the NS dataset achieved an mIoU of 0.773. However, introducing HN data into the training process improved the mIoU significantly to 0.950.

The analysis of each land cover type shows varying degrees of improvements when incorporating the HN dataset into the training. Sediment classes showed good overall improvements. For instance, the sand class, although underrepresented in the NS dataset, exhibited moderate segmentation results (0.857). With the multi-dataset training, the mIoU improved further to 0.990. The gravel class (which is also quite underrepresented in the NS dataset) improved from mIoU 0.836 to 0.971. The forest floor class had the most substantial improvement in the segmentation result, from mIoU 0.486 to 0.878.

These results suggested that multi-dataset training, even with limited sparsely annotated data, significantly enhanced the model's ability to generalize to new river environments while alleviating the class imbalance problem in the NS dataset. This can be reasoned with several factors. First, incorporating the HN dataset, where sand comprises approximately 32 \% substantially increases the model's exposure to underrepresented classes. Second, training the model on two datasets from distinct sites introduces variability in data distribution, geography, and spatial configurations. This encourages the model to learn more transferable features instead of just overfitting to site-specific features.

\subsection{Benchmarking with Other Models}

We evaluated the performance of PTv2 with the three established baselines: DGCNN \citep{phan2018dgcnn} as a pioneering deep learning network for point clouds, RandLa-Net \citep{hu2020randla} as an efficient deep learning network for point cloud applications, and RF with 500 decision trees as a traditional machine learning method. The results are illustrated in Table~\ref{tab:compare_models}. Deep learning models were trained using all features. The RF model was trained with extra geometric features, which are planarity, linearity, surface variation, and sphericity, together with all features.

\begin{table}[H]  
  \centering
  \small
  \begin{tabular}{@{}l c c  c  c  c  c  c@{}}
      \toprule
      {Model} & {\textbf{mIoU}} & {Sand} & {Gravel} & {High veg} &
      {Low veg} & {Forest floor} & {Water} \\
      \midrule     
      PTv2               & 0.950 & 0.990 & 0.971 & 0.931 & 0.930 & 0.878 & 0.998 \\
      DGCNN              & 0.744 & 0.803 & 0.331 & 0.895 & 0.682 & 0.787 & 0.967 \\
      RandLA-Net         & 0.580 & 0.597 & 0.384 & 0.897 & 0.07 & 0.544 & 0.989 \\
      RF\,(500 trees)    & 0.682 & 0.718 & 0.822 & 0.671 & 0.577 & 0.340 & 0.967 \\
      \bottomrule
  \end{tabular}
  \caption{Per-class \textbf{IoU} and \textbf{mIoU} comparing PTv2 with RF, DGCNN  and RandLA-Net.}
  \label{tab:compare_models}
\end{table}

PTv2 achieved the highest mIoU with a score of 0.950, outperforming the second-best model (DGCNN at 0.744) by 0.206 points. The performance gap varied across land cover classes, but PTv2 outperformed other methods clearly in the semantic segmentation of sediment classes like sand and gravel. All models showed robustness in water segmentation. These results highlighted PTv2's effectiveness in processing multispectral LiDAR point clouds, especially in sediment classes, where other established methods struggle to differentiate between classes.

\subsection{Predictions on the test data}
This subsection analyzes the performance of PTv2 on the JM test dataset through qualitative analysis. We present visualizations of the overall segmentation performance, discuss individual land cover classes, and identify PTv2's limitations in challenging scenarios.

\subsubsection{Overall Segmentation Performance}

\begin{figure}[H]
    \centering
    \includegraphics[width=1.1\linewidth]{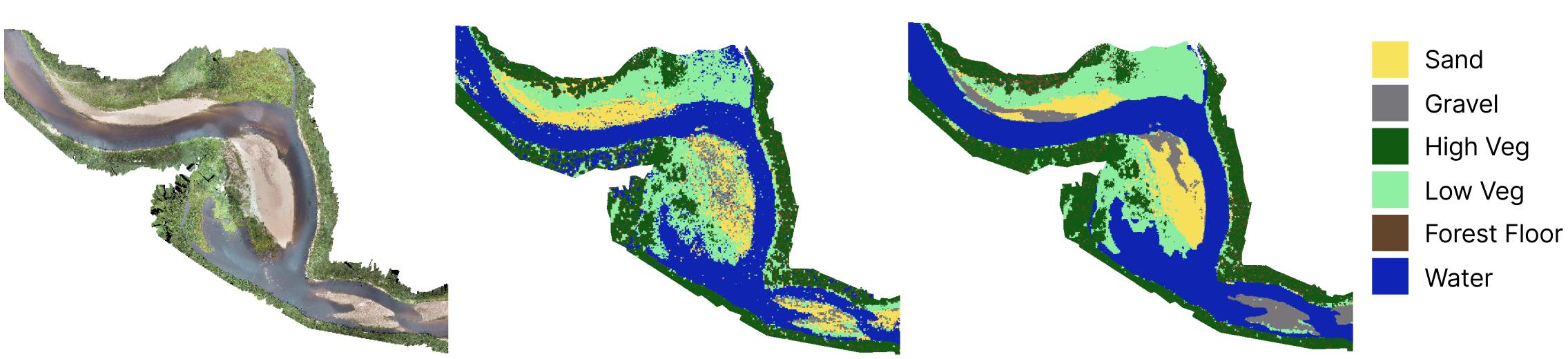}
    \caption{Orthophoto (left), prediction using only geometry (middle), and prediction using full feature set (right)}
    \label{fig:ortho_xyz_full}
\end{figure}

Figure~\ref{fig:ortho_xyz_full} presents the PTv2's semantic segmentation outputs and the corresponding orthophoto. The segmentation maps  clearly define water channels (blue) from adjacent gravel (gray) and vegetation (light green for low vegetation and dark green for high vegetation). The model successfully captures the complex morphology of the riverine environment JM. Compared to the orthophoto, the predicted boundaries align well with the visible transition.

Figure~\ref{fig:ortho_xyz_full}-middle and right compare semantic segmentation results using geometric features only (xyz) versus the full feature set. The geometry-only model produces salt and pepper noise in its predictions, especially in the terrestrial areas where sand, gravel, and low vegetation intermix. In contrast, the full feature model generates cleaner segmentation with more unified class regions and well-defined boundaries. The most substantial improvement is in sand and gravel mapping, where the spectral information proves highly beneficial for sediment mapping.

\subsubsection{Specific Land Cover Classes Discussion}

\begin{figure}[H]
    \centering
    \begin{subfigure}[t]{0.40\textwidth}
        \centering
        \includegraphics[width=0.90\linewidth]{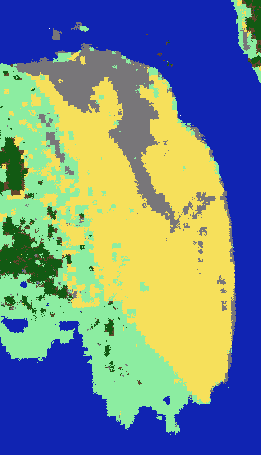}
        
        \label{fig:sub1}
    \end{subfigure}
    \begin{subfigure}[t]{0.40\textwidth}
        \centering
        \includegraphics[width=0.90\linewidth]{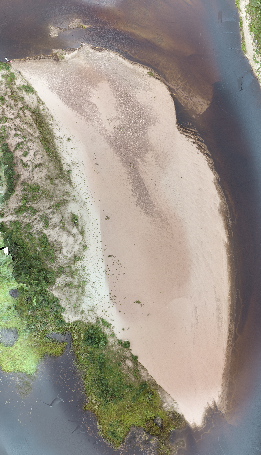}
        \label{fig:sub2}
    \end{subfigure}
    \caption{Sediment prediction results: Model prediction highlighting sediment mapping (left) and corresponding orthophoto of the study area (right)}
    \label{fig:JM_sediment}
\end{figure}

Figure~\ref{fig:JM_sediment} shows a point bar in the JM area. The orthophoto of the bar shows a majority of sandy points (lighter sediment), deposits of gravel (darker sediment), a mixture of gravel and sand, and sparse low vegetation cover. 

\begin{figure}[H]
    \centering
    \begin{subfigure}[t]{0.49\textwidth}
        \centering
        \includegraphics[width=0.95\linewidth]{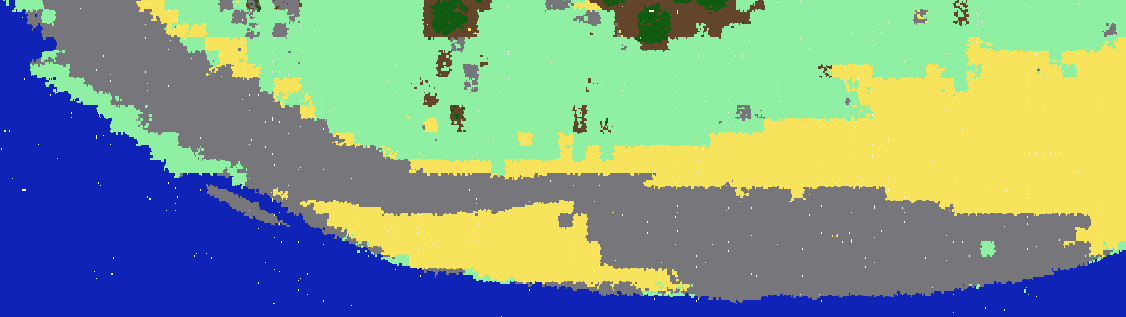}
        %\caption{Outer bank prediction}
        \label{fig:sub1}
    \end{subfigure}
    \begin{subfigure}[t]{0.49\textwidth}
        \centering
        \includegraphics[width=0.95\linewidth]{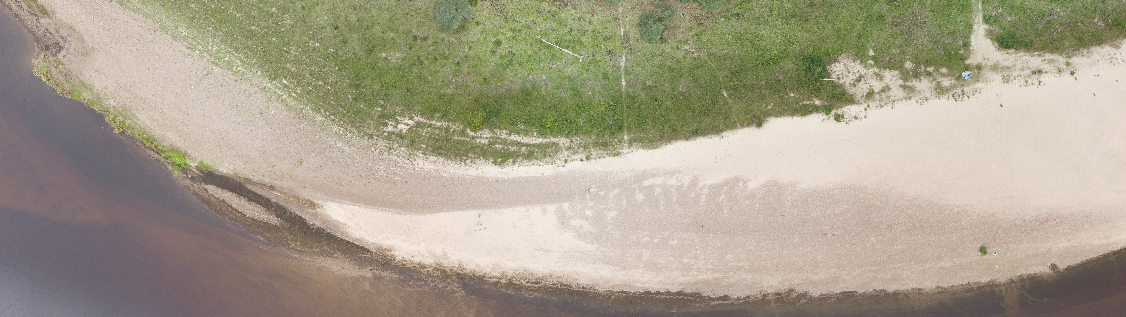}
        %\caption{Outer bank orthophoto}
        \label{fig:sub2}
    \end{subfigure}
    \caption{Outer bank mapping: prediction (left) and corresponding orthophoto (right)}
    \label{fig:outer_bank}
\end{figure}

\textbf{Sand and Gravel}- The model successfully captures the main sandy point bar with a deposit of gravel and is able to map the area where there is growth of low vegetation cover, as shown in Figure~\ref{fig:JM_sediment}. Along the outer bank in Figure~\ref{fig:outer_bank}, the model correctly maps the gravel along the river and captures the sandy area around the central area of the bank.

\begin{figure}[H]
    \centering
    \begin{subfigure}[t]{0.49\textwidth}
        \centering
        \includegraphics[width=0.95\linewidth]{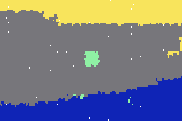}
        %\caption{Sparse vegetation prediction}
        \label{fig:sub1}
    \end{subfigure}
    \begin{subfigure}[t]{0.49\textwidth}
        \centering
        \includegraphics[width=0.95\linewidth]{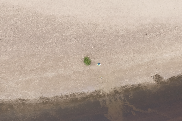}
        %\caption{Sparse vegetation orthophoto}
        \label{fig:sub2}
    \end{subfigure}
    \caption{Isolated vegetation patch mapping on the upper point bar:prediction (left) and corresponding orthophoto (right)}
    \label{fig:low_veg_sparse}
\end{figure}

\textbf{Low vegetation}- The model can map grass patches and shrubs on point bars and transitional areas, which is consistent with the low vegetation in the orthophoto as shown in~\ref{fig:JM_sediment} and Figure~\ref{fig:outer_bank}. Figure~\ref{fig:low_veg_sparse} shows that the model can map isolated vegetation cover appearing on the point bar.

\textbf{High vegetation} - High vegetation segmentation accurately identifies mature forest along the river corridor as illustrated in Figure~\ref{fig:ortho_xyz_full} (left) with the dark green color. The model can differentiate trees from the lower vegetation within the forest (forest floor).

\begin{figure}[H]
    \centering
    \begin{subfigure}[t]{0.49\textwidth}
        \centering
        \includegraphics[width=0.95\linewidth]{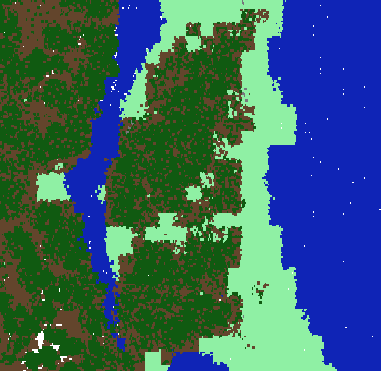}
        \label{fig:sub1}
    \end{subfigure}
    \begin{subfigure}[t]{0.49\textwidth}
        \centering
        \includegraphics[width=0.95\linewidth]{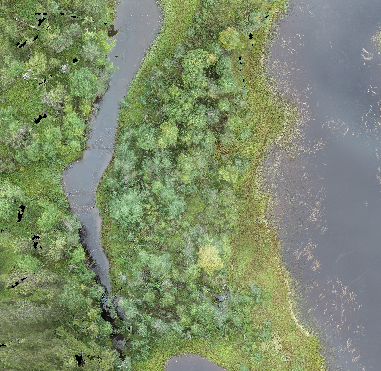}
        %\caption{}
        \label{fig:sub2}
    \end{subfigure}
    \caption{Forest floor mapping on the upper point bar : prediction (left) and corresponding orthophoto (right)}
    \label{fig:forest floor}
\end{figure}

\textbf{Forest floor} - The forest floor mapping effectively captures understorey areas beneath the forest canopy as well as areas under partial canopy cover, as shown in Figure~\ref{fig:forest floor}. The model can distinguish forest floor from sediment classes and low vegetation.

\textbf{Water} - The water class segmentation captures the river channel geometry rather accurately compared to the orthophoto, as illustrated in Figure~\ref{fig:ortho_xyz_full} (left) with the color blue. The model can map both the main river channel and minor channels and delineate water boundaries adjacent to sediment bars and vegetated banks.

\subsubsection{Challenging Scenarios}
\begin{figure}[H]
    \centering
    \includegraphics[width=0.3\linewidth]{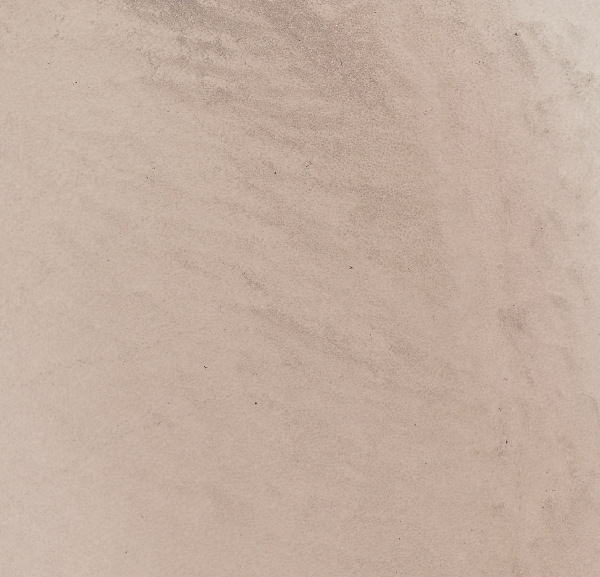}
    \caption{Sand gravel mixture area}
    \label{fig:sand-gravel mixture}
\end{figure}

\begin{figure}[H]
    \centering
    % First row - Sand gravel mixture
    \begin{subfigure}[t]{0.49\textwidth}
        \centering
        \includegraphics[width=0.95\linewidth]{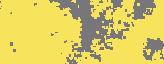}
    \end{subfigure}
    \begin{subfigure}[t]{0.49\textwidth}
        \centering
        \includegraphics[width=0.95\linewidth]{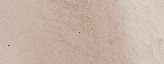}
    \end{subfigure}
    \caption*{(a)}
    \label{fig:sand gravel mix}
    
    \vspace{0.1cm}  % Vertical space between rows (adjust as needed)
    
    % Second row - Gravel float
    \begin{subfigure}[t]{0.49\textwidth}
        \centering
        \includegraphics[width=0.95\linewidth]{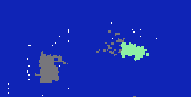}
       
    \end{subfigure}
    \begin{subfigure}[t]{0.49\textwidth}
        \centering
        \includegraphics[width=0.95\linewidth]{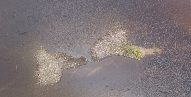}
   
    \end{subfigure}
    \caption*{(b)}
    \caption{Land cover mapping results: (a) sand-gravel mixture area and (b) gravel float mapping, with predictions (left) and corresponding orthophotos (right)}
    \label{fig:gravel_float}
\end{figure}

Sand and gravel mixtures in sediment bars or bank areas shown in Figure \ref{fig:sand-gravel mixture} can be challenging for the model to map for multiple reasons. The sediment sorting creates gradual transitions rather than sharp boundaries, making it harder for the model to predict well-defined class edges. Additionally, the grid resolution may be too coarse to capture the nuanced variations in the transitional areas. Furthermore, the model uses k-nearest neighbors to aggregate contextual information. This can be problematic in the transitional areas, as the aggregated features contain information from multiple classes. An example of model's prediction in sand-mixture zones is illustrated Figure\ref{fig:gravel_float} (a).

Figure\ref{fig:gravel_float} (b) presents another challenging mapping scenario where shallow water overlies gravel deposits (right). Some of the gravel areas and low vegetation patches are correctly classified, although the model struggles to consistently classify the whole gravel bar from the water channel. 

\section{Conclusion}

In this work, we investigated riverine land cover mapping via semantic segmentation using a transformer-based architecture deep learning model with multispectral LiDAR point cloud. 
The main findings are
\begin{itemize}
    \item The transformer-based architecture (PTv2) demonstrated strong effectiveness in land cover mapping tasks within riverine multispectral data contexts.
    \item The best performance model, evaluated by mIoU was achieved utilizing all features, which consist of 3-channel intensity, 3-channel reflectance, 3-channel amplitude, 3-channel deviation, and pulse information. 
    \item The most significant features for riverine land cover mapping are intensity and reflectance.
    \item  Multi-dataset training, even with sparse annotation from the HN dataset, was shown to significantly improve performance.
    
\end{itemize}{}

The point cloud processing pipeline is often time-consuming and computationally demanding due to the complexity and the extensive number of parameters typical in transformer-based architectures.
Despite these challenges, PTv2 has shown promising results, demonstrating the potential of transformer-based networks for semantic segmentation on multispectral LiDAR point clouds. Future works can focus on developing real-time algorithms for processing individual LiDAR sweeps in real time and optimizing the point cloud processing pipeline. To reduce reliance on annotated data, future research could also explore advanced learning frameworks such as weakly supervised, few-shot, and zero-shot learning. These frameworks are interesting because they can significantly lower annotations needed to train deep learning models, enhance models generalization in novel environments.

\section*{CRediT authorship contribution statement}
\textbf{Sopitta Thurachen:} Conceptualization, Methodology, Software, Formal analysis, Investigation, Data Curation, Writing – original draft, Writing – review and editing, Visualization, Project administration. \textbf{Josef Taher:} Methodology, Software, Writing – review and editing. \textbf{Matti Lehtomäki: }Methodology, Writing – review and editing. \textbf{Leena Matikainen: } Methodology, Writing – review and editing. \textbf{Linnea Blåfield: }Investigation, Data Curation, Writing – review and editing.  \textbf{Mikel Calle Navarro: }Investigation, Data Curation, Visualization, Writing – review and editing. \textbf{Antero Kukko:} Investigation, Resources, Writing – review and editing. \textbf{Tomi Westerlund: }Writing – review and editing, Supervision, Funding acquisition.
\textbf{Harri Kaartinen:} Conceptualization, Investigation, Resources, Writing – review and editing, Supervision, Funding acquisition.
\section*{Funding}
This work was supported by the Research Council of Finland through Digital Waters Flagship [grant number 359249, 359247], the Ministry of Education and Culture’s Doctoral Education Pilot under Decision No. VN/3137/2024-OKM-6 (Digital Waters (DIWA) Doctoral Education Pilot) and European Union - The Next Generation EU recovery instrument (RRF) through Academy of Finland projects Green-Digi-Basin [grant number 347702] and Hydrological Research Infrastructure Platform [grant number 346162].

\section*{Declaration of competing interest}
The authors declare that they have no known competing financial interests or personal relationships that could have appeared to influence the work reported in this article.

\section*{Acknowledgments}
We thank Antti Haavikko and Aurora Kauraala for their research assistance.

\bibliographystyle{elsarticle-num-names} 
\bibliography{mybibfile}

\end{document}